# Nonnegative Matrix Factorization with Zellner Penalty

**Matthew A. Corsetti, Ernest Fokoué**

School of Mathematical Sciences, Rochester Institute of Technology, Rochester, NY, USA
Email: mac4017@rit.edu, epfeqa@rit.edu

## Abstract

**Nonnegative matrix factorization (NMF) is a relatively new unsupervised learning algorithm that decomposes a nonnegative data matrix into a parts-based, lower dimensional, linear representation of the data. NMF has applications in image processing, text mining, recommendation systems and a variety of other fields. Since its inception, the NMF algorithm has been modified and explored by numerous authors. One such modification involves the addition of auxiliary constraints to the objective function of the factorization. The purpose of these auxiliary constraints is to impose task-specific penalties or restrictions on the objective function. Though many auxiliary constraints have been studied, none have made use of data-dependent penalties. In this paper, we propose Zellner nonnegative matrix factorization (ZNMF), which uses data-dependent auxiliary constraints. We assess the facial recognition performance of the ZNMF algorithm and several other well-known constrained NMF algorithms using the Cambridge ORL database.**

## 1. Introduction

Visual recognition tasks have become increasingly popular and complex in the last several decades as they often involve massively large datasets. Facial detection and recognition tasks are particularly of interest and can be severely complicated due to variation in illumination, emotional expression as well as physical location and orientation of the face within an image. Due to the often massive size of facial image datasets, subspace methods are frequently used to identify latent variables and reduce data dimensionality, so as to produce apposite representations of facial image databases.

Nonnegative matrix factorization (NMF) is a relatively new unsupervised learning subspace method that was first introduced in 1999 by Lee and Seung [1]. NMF factorizes a data matrix $X \in \mathbb{R}^{p\times n}$ while imposing a nonnegativity constraint on the matrix $X$. The subsequent nonnegative basis matrix $W \in \mathbb{R}^{p\times q}$ and nonnegative coefficient matrix $H \in \mathbb{R}^{q\times n}$ approximate $X$ when multiplied together (*i.e.* $X \approx WH$). NMF produces a sparse, part-based representation of the database as the nonnegativity constraint allows for additive, but not subtractive combinations of components. Because of this property, NMF is frequently used as a dimensionality reduction technique for tasks in which it is intuitive to combine parts to form a complete object such as in image processing, facial recognition [1]-[4] or community network visualizations [5].

Suppose $X \in \mathbb{R}^{p\times n}$ is a database of faces, for which $n$ represents the total number of images in the database and $p$ represents the number of pixels within each image (assumed to be constant across all images in the data matrix $X$). NMF factorizes the nonnegative data matrix $X$ into $W$ and $H$ by minimizing a cost function—most commonly a generalization of the square of the Euclidean Distance to matrix space

$$\left(W,H\right)=\underset{(W,H)}{\arg\min}\left\{\left\|X-WH\right\|_F^2\right\}=\underset{(W,H)}{\arg\min}\left\{\sum_{ij}\left(X_{ij}-\left(WH\right)_{ij}\right)^2\right\} \tag{1}$$

or the Kullback-Leibler Divergence

$$\left(W,H\right)=\underset{(W,H)}{\arg\min}\left\{D\left(X\,\|\,WH\right)\right\}=\underset{(W,H)}{\arg\min}\left\{\sum_{ij}\left(X_{ij}\log\frac{X_{ij}}{\left(WH\right)_{ij}}-X_{ij}+\left(WH\right)_{ij}\right)\right\}. \tag{2}$$

Many authors have adapted the NMF algorithm by altering either the cost function formulation [3] [4] [6]-[9], the minimization method for solving (1) or (2) [10]-[12], or the initialization strategy for $W$ and/or $H$ [13]-[15]. Relatively new adaptations of the NMF algorithm involve applying secondary constraints to the $W$ and/or $H$ matrix. These often take the form of smoothness constraints [8] [16] or sparsity constraints [17]-[19]. These constraints are added so as to encode prior information regarding the nature of the application under examination or to ensure preferred characteristics in the solution for $W$ and $H$. For constrained NMF (CNMF), penalty terms are used to apply the secondary constraints on $W$ and $H$. This results in an extension to the optimization task provided in (1):

$$\left(W,H\right)=\underset{(W,H)}{\arg\min}\left\{\left\|X-WH\right\|_F^2+\alpha J_1\left(W\right)+\beta J_2\left(H\right)\right\}. \tag{3}$$

Here $J_1(W)$ and $J_2(H)$ represent the penalty terms and $0\le\alpha\le1$ and $0\le\beta\le1$ are the regularization parameters that specify the relationship between the constraints. Often a sparsity constraint and an approximation error constraint are used.

Though there are many adaptations of the NMF algorithm in which auxiliary constraints are imposed on $W$ and $H$, none of these methodologies make use of data-dependent penalties. Inspired by the so-called Zellner's $g$-Prior [20], used in Bayesian Regression Analysis, we explore the use of two penalty terms that are data dependent. We use the ORL database to test the facial classification capability of the NMF algorithm when constrained by Zellner $g$-Prior penalties, henceforth referred to as Zellner nonnegative matrix factorization (ZNMF). We compare the facial classification capability of ZNMF with Constrained nonnegative matrix factorization (CNMF) [8] and show that it is superior across all selected factorization ranks. We also compare the ZNMF recognition performance with the algorithms described in [4] and determine that it outperforms many of them across many of the selected factorization ranks, most notably, the smaller of the factorization ranks.

## 2. Nonnegative Matrix Factorization Algorithms

### 2.1. Traditional Nonnegative Matrix Factorization with Multiplicative Update

NMF factorizes a matrix $X \in \mathbb{R}^{p\times n}$ into a basis matrix $W \in \mathbb{R}^{p\times q}$ and a coefficient matrix $H \in \mathbb{R}^{q\times n}$ while imposing a nonnegativity constraint. Because of the nonnegativity constraint, the basis images (when considering $X$ to be a database of faces) can be combined in an additive fashion to form a complete face. In traditional NMF [1] the two most commonly considered cost functions for determining the cost of factorizing $X$ into $W$ and $H$ are the square of the Euclidean (1) and the Kullback-Leibler Divergence (2). Traditional NMF produces the $W$ and $H$ matrices by calculating minimizations of (1) or (2) using the following multiplicative update equations:

$$W^{(t+1)} = W^{(t)} \left[ \frac{XH^{(t)\mathrm{T}}}{W^{(t)}H^{(t)}H^{(t)\mathrm{T}}} \right] = W^{(t)} \eta\left(W^{(t)}\right) \tag{4}$$

where

$$\eta\left(W^{(t)}\right) = \frac{X\left(H^{(t)}\right)^{\mathrm{T}}}{W^{(t)}H^{(t)}\left(H^{(t)}\right)^{\mathrm{T}}} \tag{5}$$

and

$$H^{(t+1)} = H^{(t)} \left[ \frac{W^{(t+1)\mathrm{T}}X}{W^{(t+1)\mathrm{T}}W^{(t+1)}H^{(t)}} \right] = H^{(t)} \psi\left(H^{(t)}\right) \tag{6}$$

where

$$\psi\left(H^{(t)}\right) = \frac{\left(W^{(t+1)}\right)^{\mathrm{T}} X}{\left(W^{(t+1)}\right)^{\mathrm{T}} W^{(t+1)} H^{\mathrm{T}}}. \tag{7}$$

Traditional NMF using a multiplicative update is known to be slow to converge as it requires a large number of iterations. Gradient descent and alternating least squares algorithms are commonly used in place of traditional NMF as they require far fewer iterations resulting in a faster convergence; however, we will not explore them in this paper.

The standard NMF multiplicative updating algorithms have a continuous descent property. The descent will lead to a stationary point within the region under examination; however, it is uncertain as to whether or not this stationary point is a local minimum as it could certainly be a saddle point. This is due to the iterative optimization nature of the algorithm which optimizes $W$ and $H$ iteratively, though never simultaneously.

### 2.2. Constrained Nonnegative Matrix Factorization

CNMF [8] expands the optimization task shown in (1) to include penalty terms $J_1(W)$ and $J_2(H)$ that serve to apply task-specific, auxiliary constraints on the solutions of (3). $0 \le \alpha \le 1$, $0 \le \beta \le 1$ are regularization parameters. For our purposes we define $J_1(W)$ and $J_2(H)$ from (3) as follows:

$$J_1(W) = \|W\|_F^2 \tag{8}$$

and

$$J_2(H) = \|H\|_F^2, \tag{9}$$

while $\alpha$ and $\beta$ are used as constraints on the sparsity and approximation error respectively. When the optimization task is that of (3), the multiplicative updates of (4) and (6) are modified as follows:

$$W^{(t+1)} = W^{(t)} \frac{X\left(H^{(t)}\right)^{\mathrm{T}}}{W^{(t)}H^{(t)}\left(H^{(t)}\right)^{\mathrm{T}} + \alpha \frac{\partial}{\partial W} J_1\left(W^{(t)}\right)} \tag{10}$$

and

$$H^{(t+1)} = H^{(t)} \frac{\left(W^{(t+1)}\right)^{\mathrm{T}} X}{\left(W^{(t+1)}\right)^{\mathrm{T}} W^{(t+1)} H^{(t)} + \beta \frac{\partial}{\partial H} J_2\left(H^{(t)}\right)}. \tag{11}$$

### 2.3. Zellner Nonnegative Matrix Factorization

In regression analysis, for a Gaussian Distribution with $p\left(Y \mid \beta, X, \sigma^2\right) = N\left(X\beta, \sigma^2 I\right)$ it is known that

$$\hat{\beta}^{(OLS)} = \left(X^{\mathrm{T}} X\right)^{-1} X^{\mathrm{T}} Y \tag{12}$$

has variance

$$V\left(\hat{\beta}\right) = \sigma^2 \left(X^{\mathrm{T}} X\right)^{-1}. \tag{13}$$

The Zellner $g$-Prior exploits this fact, in the Bayesian setting to use

$$p\left(\beta \mid g, \sigma^2\right) = N\left(0, g\sigma^2 \left(X^{\mathrm{T}} X\right)^{-1}\right) \tag{14}$$

which corresponds to using the penalty's empirical risk shown below:

$$\mathcal{E}(\beta) = \|Y - X\beta\|_F^2 + \frac{1}{g}\beta^{\mathrm{T}}\left(X^{\mathrm{T}} X\right)\beta$$

$$\mathcal{E}(\beta) = (Y - X\beta)^{\mathrm{T}}(Y - X\beta) + \frac{1}{g}\beta^{\mathrm{T}}\left(X^{\mathrm{T}} X\right)\beta$$

$$\frac{\partial \mathcal{E}(\beta)}{\partial \beta} = -2X^{\mathrm{T}}(Y - X\beta) + \frac{2}{g}\left(X^{\mathrm{T}} X\right)\beta$$

$$\frac{\partial \mathcal{E}(\beta)}{\partial \beta} = 0 \Leftrightarrow -X^{\mathrm{T}} Y + X^{\mathrm{T}} X\beta + \frac{1}{g}\left(X^{\mathrm{T}} X\right)\beta = 0$$

$$\Leftrightarrow \left(1 + \frac{1}{g}\right) X^{\mathrm{T}} X\beta = X^{\mathrm{T}} Y$$

$$\Leftrightarrow \hat{\beta} = \left(\frac{g}{1+g}\right)\left(X^{\mathrm{T}} X\right)^{-1} X^{\mathrm{T}} Y.$$

We extend and adapt Zellner's ideas as follows:

$$J_1(W) = \mathrm{trace}\left(\underset{q\times p}{W^{\mathrm{T}}} \underset{p\times n}{X} \underset{n\times p}{X^{\mathrm{T}}} \underset{p\times q}{W}\right) = \mathrm{trace}\left(S^{\mathrm{T}} S\right) \tag{15}$$

where $S = (X^{\mathrm{T}}W)$ is $n \times q$, $S^{\mathrm{T}}$ is $q \times n$ and $S^{\mathrm{T}}S$ is $q \times q$.

$$J_2(H) = \mathrm{trace}\left(HX^{\mathrm{T}} XH^{\mathrm{T}}\right) = \mathrm{trace}\left(R^{\mathrm{T}} R\right) \tag{16}$$

where $R = XH^{\mathrm{T}}$ is $p \times q$ and essentially represents the projection weighting. $R^{\mathrm{T}}R$ is $q \times q$ and its diagonal essentially represents the idiosyncratic variance of the projections onto the lower dimensional space.

$$\frac{\partial}{\partial W} J_1(W) = \frac{\partial}{\partial W}\mathrm{trace}\left(W^{\mathrm{T}} XX^{\mathrm{T}} W\right) = XX^{\mathrm{T}} W + XX^{\mathrm{T}} W = 2XX^{\mathrm{T}} W \tag{17}$$

$$\frac{\partial}{\partial H} J_2(H) = \frac{\partial}{\partial H}\mathrm{trace}\left(HX^{\mathrm{T}} XH^{\mathrm{T}}\right) = HX^{\mathrm{T}} X + HX^{\mathrm{T}} X = 2HX^{\mathrm{T}} X \tag{18}$$

As can be seen in (17) and (18), the updates of both $W$ and $H$ are simply post or pre-weighted by the input space variances or the data spaces variances.

Our objective function is

$$F(W, H) = \|X - WH\|_F^2 + \alpha J_1(W) + \beta J_2(H) \tag{19}$$

where

$$J_1(W) = \frac{1}{g}\mathrm{trace}\left(W^{\mathrm{T}} XX^{\mathrm{T}} W\right), \tag{20}$$

and

$$J_2(H) = \frac{1}{g}\mathrm{trace}\left(HX^{\mathrm{T}} XH^{\mathrm{T}}\right), \tag{21}$$

where

$$g = \max\left(n, p^2\right). \tag{22}$$

For (22) the Risk Inflation Criterion (RIC) of Foster and George [21], which sets $g = p^2$ is combined with the Bayesian Information Criterion (BIC) to produce the so-called benchmark prior as $g = n$ leads to the unit information prior found in BIC. And so, (22) will be found to be appropriate.

When using ZNMF, the updating equations of CNMF shown in (10) and (11) are modified as follows:

$$W^{(t+1)} = W^{(t)} \frac{X\left(H^{(t)}\right)^{\mathrm{T}}}{W^{(t)}H^{(t)}\left(H^{(t)}\right)^{\mathrm{T}} + \left(\frac{\alpha}{g}\right)XX^{\mathrm{T}}W^{(t)}} \tag{23}$$

and

$$H^{(t+1)} = gH^{(t)}\left[\frac{\left(W^{(t+1)}\right)^{\mathrm{T}} X}{g\left(W^{(t+1)}\right)^{\mathrm{T}} W^{(t+1)}H^{(t)} + \beta H^{(t)}X^{\mathrm{T}}X}\right]. \tag{24}$$

## 3. Experimental Results

In this section, we conducted a series of simulations to evaluate the classification performance of the ZNMF and CNMF algorithms. We replicated the ORL classification experiment conducted in Wang *et al*. [4], which evaluated the classification performance of traditional NMF, Local NMF (LNMF) [6], Fisher NMF (FNMF) [4], Principle Component Analysis, and Principle Component Analysis NMF (PNMF) [4] using the Cambridge ORL database. By replicating the experiment in [4], many hundreds of times, we created an avenue through which direct comparisons of the performances of the aforementioned algorithms could be carried out.

The Cambridge ORL database consists of 10 gray-scale facial images each of 36 male and 4 female subjects. The images vary in illumination, facial expression and position. The faces are forward-facing with slight rotations to the left and right. For each simulation, the training dataset $X \in \mathbb{R}^{644\times 200}$ was produced by randomly selecting 5 images from each of the 40 subjects resulting in a training dataset of 200 images of 644 pixels each. The test datasets were comprised of the remaining 200 unselected images and were used to evaluate the facial recognition capabilities of CNMF and ZNMF using the first Nearest Neighbor classifier. In order to optimize the computational efficiency the resolution of the images was reduced from 112 × 92 to 28 × 23 in accordance with [4], which found that reducing the resolution of the ORL faces to 25% of the original resolution had little effect on the accuracy of the facial recognition. The reduction in resolution is demonstrated for 9 images shown in **Figure 1**.

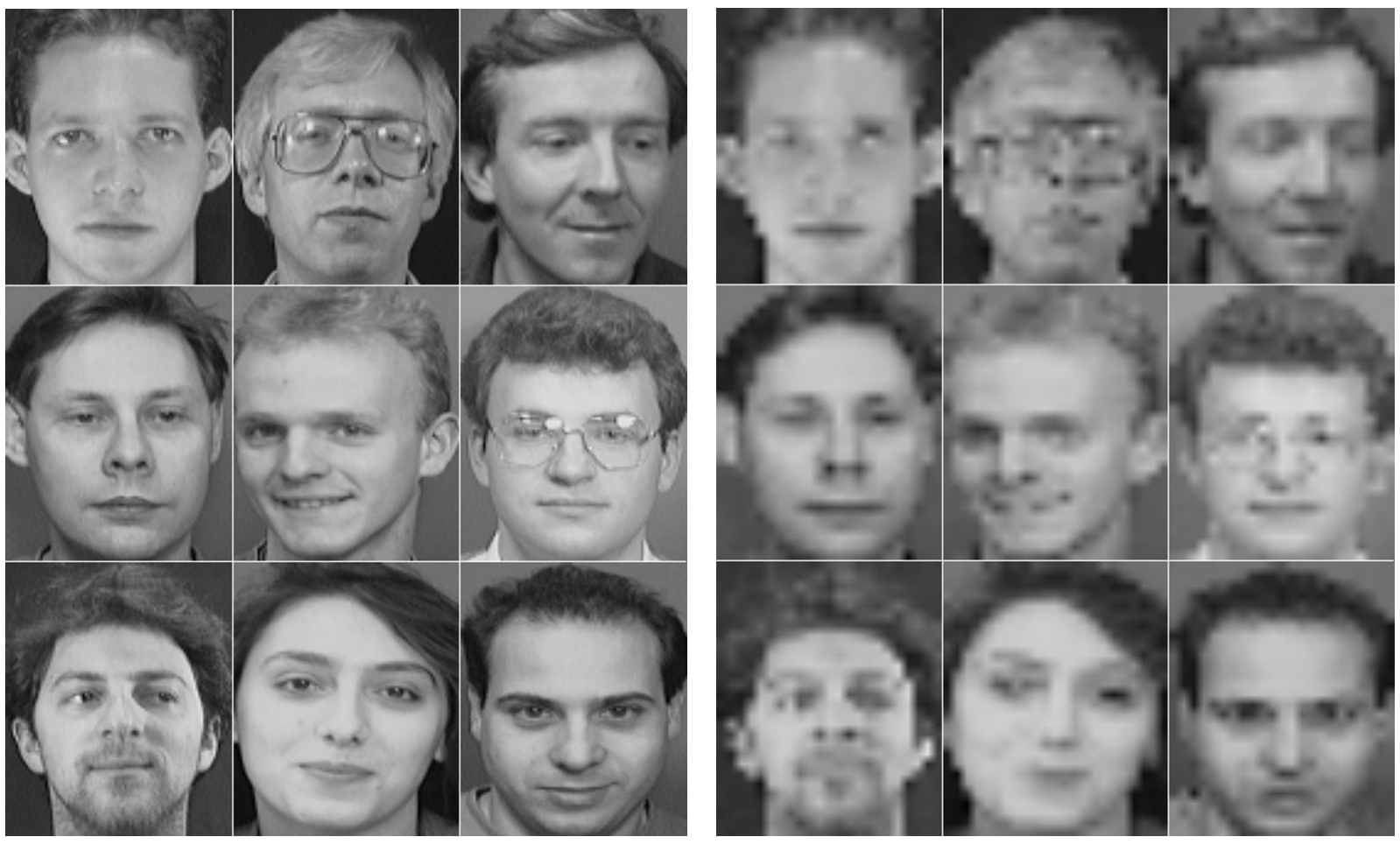

**Figure 1.** (Left) 9 ORL faces at full 112 × 92 resolution; (Right) 9 ORL faces at reduced 28 × 23 resolution.

The effects of the $\alpha$, $\beta$, and $g$-Prior parameter settings on the average recognition rate were explored to great lengths through extensive computer simulations. We restricted the $\alpha$, $\beta$ relationship to two possible scenarios:

$$\alpha = \beta \tag{25}$$

and

$$\alpha = 1 - \beta, \tag{26}$$

such that $0 \le \alpha \le 1$, $0 \le \beta \le 1$. Optimal $\alpha$ and $\beta$ settings were determined across all considered factorization ranks $q \in \{16, 25, 36, 49, 64, 81, 100\}$ for the CNMF algorithm simulations using (25) and (26) (see **Table 1** and **Table 2**). This was not the case for the ZNMF simulations because of the addition of the $g$-Prior parameter which dramatically increased the number of possible settings for the regularization parameters. Because there were many more regularization parameter settings to consider for the ZNMF algorithm than the CNMF algorithm, $\alpha$ and $\beta$ were optimized exclusively at a factorization rank $q$ = 16 for (25) and (26) in the ZNMF simulations. The optimal tuning values of $\alpha$ and $\beta$ for the ZNMF simulations, were then held constant across the remaining factorization ranks (25, 36, 49, 64, 81, 100), only differing depending upon the relationship of $\alpha$ with $\beta$ specified by (25) and (26). There were 20 replications used at each unique setting of the regularization parameters for the CNMF algorithm; while only 5 replications were used at each unique setting of the regularization parameters in the ZNMF simulations. The noticeable difference between the number of replications for the CNMF and ZNMF algorithms was again due to the fact that there were far more parameter settings to explore using ZNMF than CNMF.

**Table 1.** CNMF optimized regularization parameter settings and recognition performances using the $\alpha$ $\beta$ relationships of (25) and (26).

| Factorization Rank (q) | $\alpha$ and $\beta$ Relationship | Optimal $\alpha$ | Optimal $\beta$ | Average Recognition Rate |
|---|---|---|---|---|
| 16 | $\alpha = \beta$ | 0.81 | 0.81 | 0.88578 |
| | $\alpha = 1 - \beta$ | 0.15 | 0.85 | 0.88602 |
| 25 | $\alpha = \beta$ | 0.64 | 0.64 | 0.88631 |
| | $\alpha = 1 - \beta$ | 0.39 | 0.61 | 0.88698 |
| 36 | $\alpha = \beta$ | 0.56 | 0.56 | 0.88630 |
| | $\alpha = 1 - \beta$ | 0.82 | 0.18 | 0.88347 |
| 49 | $\alpha = \beta$ | 0.99 | 0.99 | 0.88685 |
| | $\alpha = 1 - \beta$ | 0.82 | 0.18 | 0.88556 |
| 64 | $\alpha = \beta$ | 0.42 | 0.42 | 0.88838 |
| | $\alpha = 1 - \beta$ | 0.46 | 0.54 | 0.88561 |
| 81 | $\alpha = \beta$ | 1.00 | 1.00 | 0.88868 |
| | $\alpha = 1 - \beta$ | 0.14 | 0.86 | 0.88460 |
| 100 | $\alpha = \beta$ | 0.94 | 0.94 | 0.88901 |
| | $\alpha = 1 - \beta$ | 0.12 | 0.88 | 0.88496 |

**Table 2.** ZNMF optimized regularization parameter settings and recognition performances using the $\alpha$ $\beta$ relationships of (25) and (26).

| Factorization Rank ($q$) | $\alpha$ and $\beta$ Relationship | Optimal $\alpha$ | Optimal $\beta$ | Optimal $g$-Prior | Average Recognition Rate |
|---|---|---|---|---|---|
| 16 | $\alpha = \beta$ | 0.45 | 0.45 | 75 | 0.90283 |
| | $\alpha = 1 - \beta$ | 0.60 | 0.40 | 80 | 0.90080 |
| 25 | $\alpha = \beta$ | 0.45 | 0.45 | 75 | 0.89975 |
| | $\alpha = 1 - \beta$ | 0.60 | 0.40 | 80 | 0.90200 |
| 36 | $\alpha = \beta$ | 0.45 | 0.45 | 75 | 0.90039 |
| | $\alpha = 1 - \beta$ | 0.60 | 0.40 | 80 | 0.89933 |
| 49 | $\alpha = \beta$ | 0.45 | 0.45 | 75 | 0.90093 |
| | $\alpha = 1 - \beta$ | 0.60 | 0.40 | 80 | 0.89952 |
| 64 | $\alpha = \beta$ | 0.45 | 0.45 | 75 | 0.90055 |
| | $\alpha = 1 - \beta$ | 0.60 | 0.40 | 80 | 0.90219 |
| 81 | $\alpha = \beta$ | 0.45 | 0.45 | 75 | 0.90024 |
| | $\alpha = 1 - \beta$ | 0.60 | 0.40 | 80 | 0.89918 |
| 100 | $\alpha = \beta$ | 0.45 | 0.45 | 75 | 0.90111 |
| | $\alpha = 1 - \beta$ | 0.60 | 0.40 | 80 | 0.90012 |

The recognition performances of the ZNMF simulations across various settings of $\alpha$, $\beta$, and the $g$-Prior are on display in **Figure 2** and **Figure 3**. We were able to determine the optimal settings for $\alpha$, $\beta$, and the $g$-Prior for the ZNMF simulations using these surfaces. Initially we explored two broad regions. The first, shown to the left in **Figure 2**, took into consideration the regularization parameter relationship specified by (25); while the second, shown in the right of **Figure 2** considered the relationship specified by (26). Both surfaces in **Figure 2** depict a maximal region defined by relatively low *g*-Prior values and $0.40 \le \alpha \le 0.60$. The natures of these optimal regions were explored using the surfaces of **Figure 3**. There were 25 replications conducted at each of the unique regularization parameter settings in these condensed territories. The optimal parameter settings for the ZNMF algorithm under both condition (25) and condition (26), using a factorization rank $q = 16$, were discovered atop ridgelines in the optimal territories of **Figure 3** and are provided in **Table 2**.

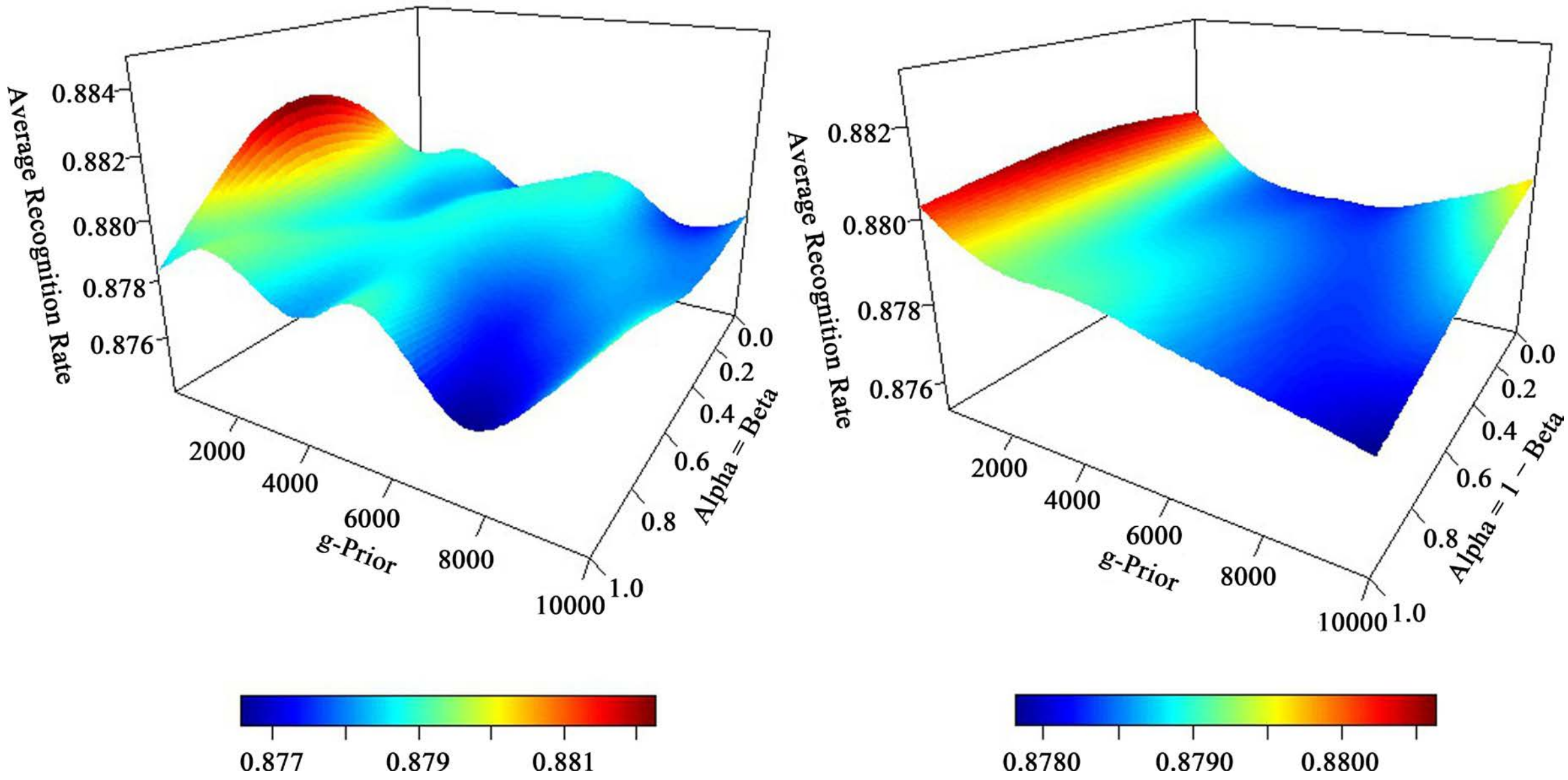

**Figure 2.** (Left) Recognition performances of ZNMF with $\alpha = \beta$; (Right) Recognition performances of ZNMF with $\alpha = 1 - \beta$.

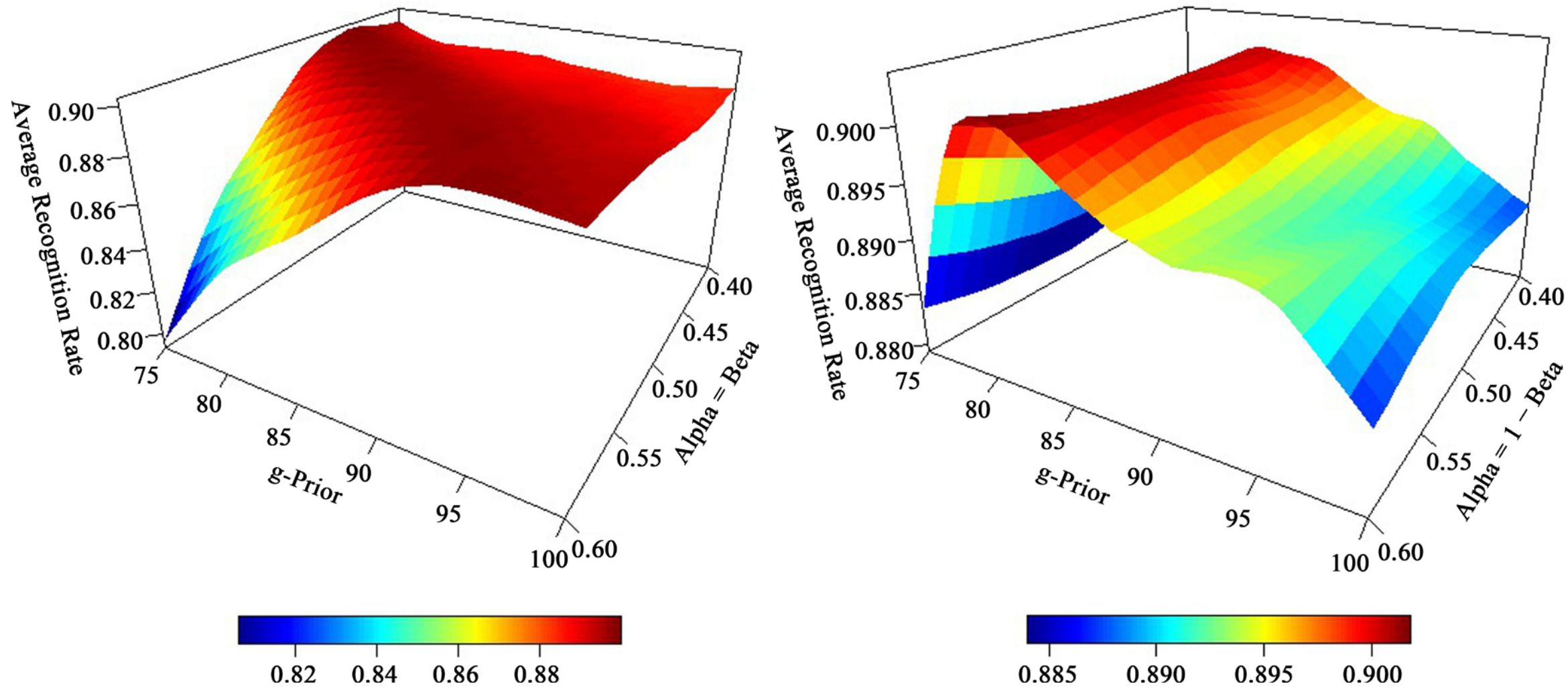

**Figure 3.** (Left) Recognition performances of ZNMF with $\alpha = \beta$ in optimal condensed territory; (Right) Recognition performances of ZNMF with $\alpha = 1 - \beta$ in optimal condensed territory.

After identifying optimal settings for the regularization parameters, 500 replications were conducted for both the CNMF and ZNMF algorithms, across the factorization ranks $q \in \{16, 25, 36, 49, 64, 81, 100\}$, using the optimal parameter settings. The results, provided in **Figure 4** were quite telling. The ZNMF algorithm had a better average recognition rate across all factorization ranks for both (25) and (26) than the CNMF algorithm. Furthermore, the ZNMF algorithm produced better average recognition rates than the NMF, LNMF, FNMF, PCA and PNMF algorithms used in [4] across the majority of the factorization ranks. The first exception to this occurred at factorization rank of $q = 49$, in which ZNMF performed better than NMF, LNMF, and PCA, and approximately equal to FNMF and PNMF. The second and third exceptions occurred at factorization ranks $q = 64$ and $q = 81$ where ZNMF outperformed NMF, LNMF, PCA and PNMF and performed approximately equal to FNMF. It should be noted that the ZNMF algorithm was able to maintain relatively higher recognition rates (about 90%) consistently across all factorization ranks, including smaller factorization ranks, such as $q = 16$ and $q = 25$ where other algorithms produced lower average recognition rates. This is quite exciting as it implies that ZNMF requires less information (lowered factorization ranks) to produce equally as impressive recognition rates on the ORL database as other algorithms [4] produce when provided with relatively more information (higher factorization ranks).

## 4. Conclusion and Discussion

In this paper, we proposed the ZNMF algorithm for the assessment of facial recognition and assessed its capability in this regard using the Cambridge ORL Faces Database. We compared its facial recognition capabilities with traditional NMF and several constrained version of NMF across seven different factorization ranks. We found that ZNMF algorithm outperformed the other algorithms across the majority of the factorization ranks, most notably at the lower factorization ranks where the margin of improvement was the most significant. The FNMF algorithm approximately tied the facial recognition rate of the ZNMF algorithm at three factorization ranks (49, 64 and 81) and the PNMF algorithm approximately tied the ZNMF algorithm at just one factorization rank (49). Quite possibly the most important finding was that the ZNMF algorithm produced facial recognition rates, using less information (lower factorization ranks), that either out-performed or were comparable to the results of other algorithms at higher factorization ranks. This finding implied that, for the ORL Dataset, the data-dependent ZNMF algorithm could classify facial images better than the other algorithms under examination and it could do so with less information, making it computationally less taxing.

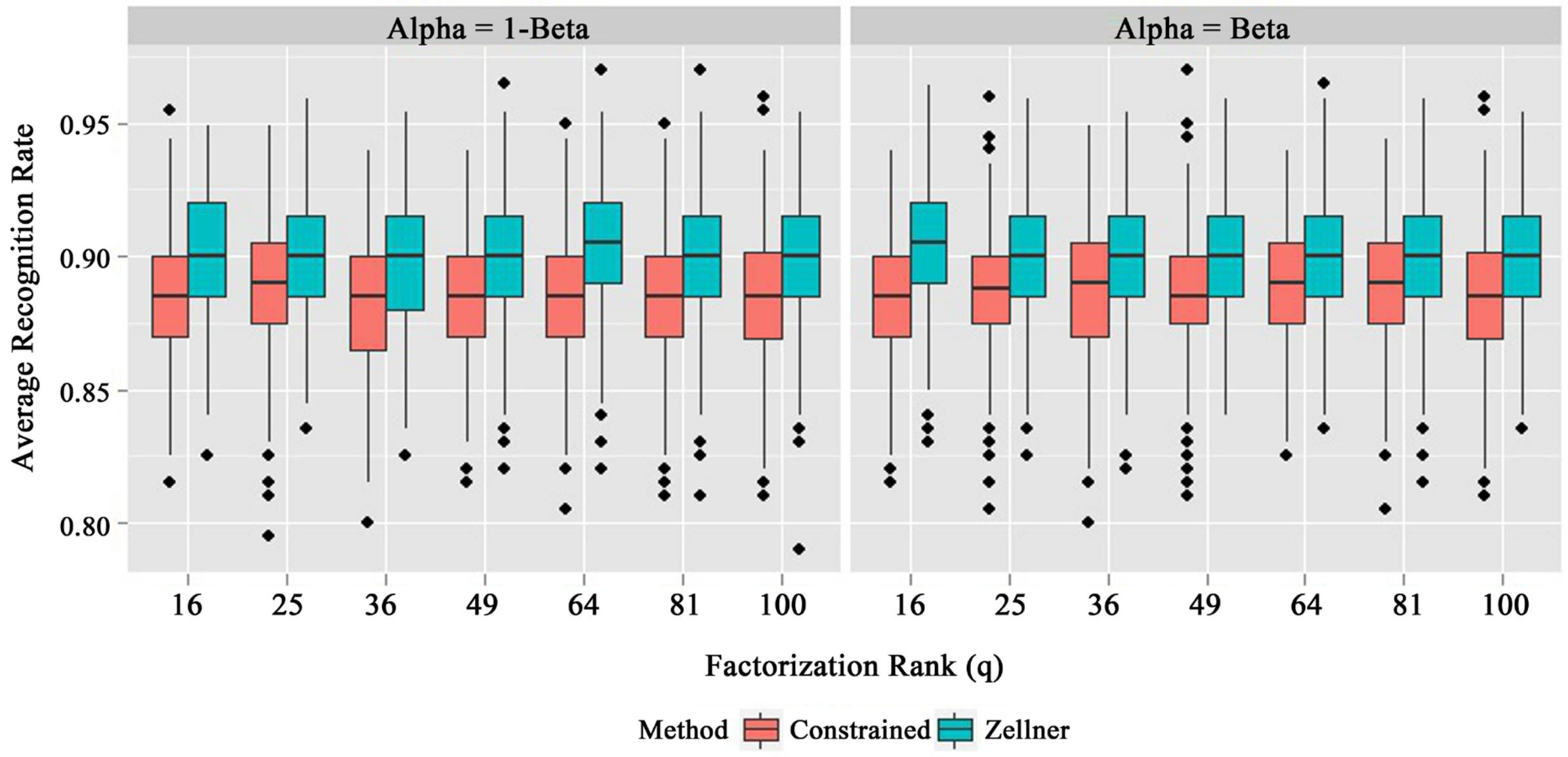

**Figure 4.** Average correct recognition rates of the CNMF and ZNMF algorithms using the ORL database with 500 simulations at each factorization rank $q \in \{16, 25, 36, 49, 64, 81, 100\}$. $q$ was determined in accordance with [4].

This paper demonstrates the advantages of including data-dependent auxiliary constraints in the NMF algorithm through the introduction of ZNMF. In the future, we hope to explore other data-dependent auxiliary constraints. A possibility would be to use $G_1(W) = \text{trace}(X^{\mathrm{T}}WW^{\mathrm{T}}) = \text{trace}(SS^{\mathrm{T}})$ where $S = X^{\mathrm{T}}W \in \mathbb{R}^{n\times q}$, and $G_2(H) = \text{trace}(XH^{\mathrm{T}}HX^{\mathrm{T}}) = \text{trace}(RR^{\mathrm{T}})$ where $R = XH^{\mathrm{T}} \in \mathbb{R}^{p\times q}$. Notice that $SS^{\mathrm{T}}$ is $n \times n$ and is somewhat the linear Gram matrix of the projected $X$. $RR^{\mathrm{T}}$ is $p \times p$ and somewhat mirrors the covariance matrix in input space. We hope to explore these auxiliary constraints in the near future, again using the Cambridge ORL database and perhaps the Facial Recognition Technology (FERET) database as well.

## Acknowledgements

Ernest Fokoué wishes to express his heartfelt gratitude and infinite thanks to our lady of perpetual help for her ever-present support and guidance, especially for the uninterrupted flow of inspiration received through her most powerful intercession.